\documentclass[10pt,twocolumn,letterpaper]{article}

\usepackage{CJKutf8}
\usepackage{bm}

\usepackage{iccv}
\usepackage{times}
\usepackage{epsfig}
\usepackage{graphicx}
\usepackage{amsmath}
\usepackage{amssymb}

\usepackage{algorithm}
\usepackage{algpseudocode}
\usepackage{comment}
\usepackage[toc,page]{appendix}
\usepackage{here}

\usepackage[breaklinks=true,bookmarks=false]{hyperref}
\usepackage[font=small,labelfont=bf,tableposition=top]{caption}

\iccvfinalcopy 

\def\iccvPaperID{6854} 

\ificcvfinal\fi

\begin{document}

\title{Instruct 3D-to-3D: Text Instruction Guided 3D-to-3D conversion}

\def\correspondingauthor{\footnote{Corresponding author: Hiromichi.Kamata@sony.com}}

\author{Hiromichi Kamata\footnotemark \quad Yuiko Sakuma \quad Akio Hayakawa \quad Masato Ishii \quad Takuya Narihira\\
Sony Group Corporation\\
Tokyo, Japan\\
{\tt\small \{Hiromichi.Kamata, Yuiko.Sakuma, Akio.Hayakawa, Masato.A.Ishii, Takuya.Narihira\}@sony.com}
}

\makeatletter
\let\@oldmaketitle\@maketitle
\renewcommand{\@maketitle}{\@oldmaketitle
\begin{center}
  \includegraphics[width=\textwidth]
    {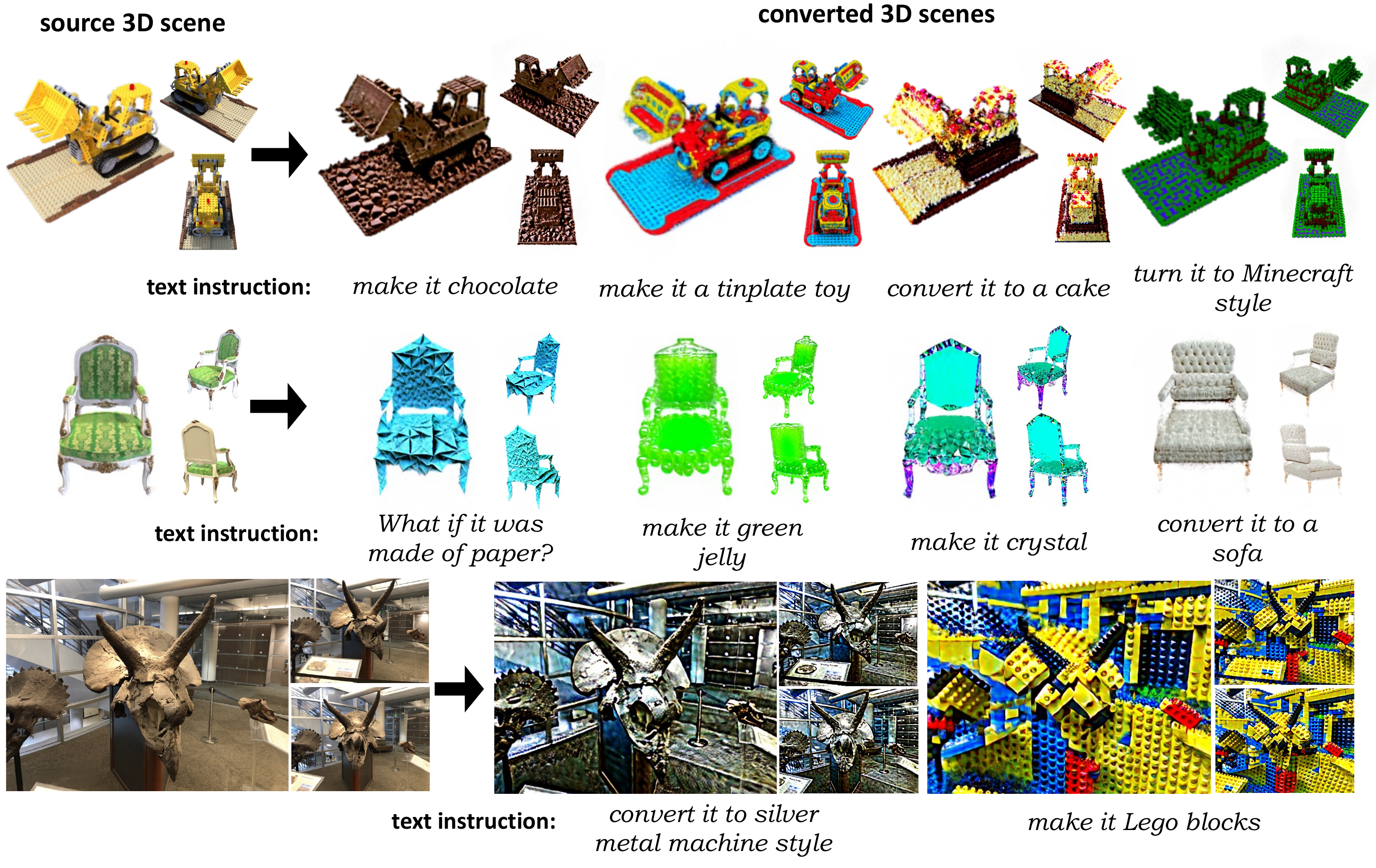}
    \captionof{figure}{Instruct 3D-to-3D Results. Our Instruct 3D-to-3D successfully converts input 3D scenes according to the text instructions.}
    \label{fig:top}
    \bigskip
\end{center}
}
    
\makeatother

\maketitle
\ificcvfinal\thispagestyle{empty}\fi

\renewcommand{\thefootnote}{\fnsymbol{footnote}}	
\renewcommand\thefootnote{*}
\footnotetext{Corresponding author: Hiromichi.Kamata@sony.com}

\begin{CJK}{UTF8}{ipxm}

\begin{abstract}
We propose a high-quality 3D-to-3D conversion method, Instruct 3D-to-3D. Our method is designed for a novel task, which is to convert a given 3D scene to another scene according to text instructions. Instruct 3D-to-3D applies pretrained Image-to-Image diffusion models for 3D-to-3D conversion. This enables the likelihood maximization of each viewpoint image and high-quality 3D generation. In addition, our proposed method explicitly inputs the source 3D scene as a condition, which enhances 3D consistency and controllability of how much of the source 3D scene structure is reflected. We also propose dynamic scaling, which allows the intensity of the geometry transformation to be adjusted.
We performed quantitative and qualitative evaluations and showed that our proposed method achieves higher quality 3D-to-3D conversions than baseline methods. \url{https://sony.github.io/Instruct3Dto3D-doc/}
\end{abstract}

\section{Introduction}

In recent years, generative modelings based on diffusion models have been rapidly developed and applied in a wide range of domains, including image \cite{dalle2,imagen, sd}, music \cite{noise2music,diffmusic}, video \cite{makeavideo,imagenvideo}, and 3D \cite{dreamfusion,magic3d}. 
Especially for Text-to-Image (T2I) tasks, diffusion models have received much attention for their ability to generate high-quality images that align with input texts. Such amazing T2I models are often realized by large-scale training with a huge amount of image-text pair dataset \cite{dalle2, imagen, sd}.

The pretrained T2I models can be also utilized to solve various tasks other than T2I.
In particular, the application of diffusion models to the field of 3D has been studied extensively \cite{dreamfusion, magic3d, latentnerf, sjc}. They used the pretrained T2I models to optimize NeRF \cite{nerf} so that the content of rendered images from NeRF always matches input texts. Through this optimization, they successfully generate NeRF representation of the 3D scene described by the input texts.

Nowadays, it has become much easier to obtain NeRF representations of real-world scenes \cite{nerf, barron2021mipnerf}. If these 3D scenes can be edited with text instructions, it should also make it substantially easier to create high-quality 3D content. Prior works \cite{clipnerf,nerfart} have shown that it is possible to stylize a 3D scene based on a given text that describes the target scene to be obtained after the editing. However, in their methods, we cannot directly instruct what scene to be changed, which fairly reduces the controllability of the editing. In addition, we cannot also accurately control how strongly the structure of the original scene would be preserved after the editing.

In this paper, we tackle a novel and challenging problem, which is to convert a given 3D scene to another scene according to text instructions. To solve this problem, we propose \textbf{Instruct 3D-to-3D}, which performs 3D-to-3D conversions using pretrained diffusion models.

Our Instruct 3D-to-3D achieves better 3D consistency, quality, and controllability simultaneously compared to baseline methods.
For better 3D consistency, we directly use the source 3D scene as a condition to keep its semantic and structural information.
Furthermore, to achieve better quality, we apply a pretrained Image-to-Image (I2I) diffusion model that allows us to maximize the likelihood of each viewpoint image.
In addition, we propose dynamic scaling to enable the users to control the strength of the geometry transformation.
We use voxel grid-based implicit 3D representation \cite{dvgo,plenoxels} as a 3D model. Our dynamic scaling gradually decreases and increases the 3D resolution of the voxel grid to achieve controllable and smooth 3D geometry conversions.

The main contributions of this study are summarized as follows.
\begin{itemize}
    \item We propose Instruct 3D-to-3D, a method for transforming a 3D scene based on text instructions. Our Instruct 3D-to-3D realizes the high quality and 3D-consistent conversion using a pretrained I2I model conditioned by the source 3D scene.
    \item We propose dynamic scaling. This allows manipulation of the strength of the geometry transformation and smooth 3D-to-3D conversion.
    \item We conducted qualitative and quantitative evaluations, showing that our proposed method can perform 3D-to-3D conversion with higher quality than baseline methods.
\end{itemize}

\section{Related Works}

\subsection{Diffusion Models}
Diffusion models \cite{diffusion2015,beatgans} are generative models inspired by non-equilibrium thermodynamics. It can generate high-fidelity images by gradually denoising data starting from pure random noise.
The diffusion models comprise two processes along with timesteps: a forward process where a small amount of noise is added to the data at each timestep, and a reverse process where the data are slightly denoised at each timestep. Specifically, in the forward process, the data at timestep $t$ can be obtained as follows:

\begin{align}
    \bm{x}_t &= \sqrt{\alpha_t}\bm{x}_{t-1} + \sqrt{1-\alpha_t}\bm{\epsilon}_{t-1} \nonumber\\
     &= \cdots = \sqrt{\bar{\alpha}_t}\bm{x}_0 + \sqrt{1-\bar{\alpha}_t} \bm{\epsilon}
\end{align}
Here, $\{\alpha_i \}_{i=1}^T$ are hyper-parameters that satisfy $0 < \alpha_i < 1$ and $\bar{\alpha}_t = \Pi_{i=1}^t \alpha_i$, $0<\bar{\alpha}_T < \bar{\alpha}_{T-1}< \cdots < \bar{\alpha}_T < 1 $. $\bm{\epsilon}$ is a random noise sampled as $\bm{\epsilon} \sim \mathcal{N}(0, \bm{I})$.

In the reverse process, The noise added at each timestep needs to be predicted from noisy data using a neural network to denoise the data. We may optionally use condition information for this prediction, and here we use an input text $y$. The predicted noise is denoted as $\epsilon_\phi (\bm{x}_t; y, t)$, where $\phi$ is the parameters of the neural network. The neural network is trained to minimize the following loss function.
\begin{align}
    \mathcal{L} = \mathbb{E}_{t,\bm{\epsilon}} \left[ w(t)\|\epsilon_\phi (\bm{x}_t; y, t) - \bm{\epsilon}\|^2 \right]
\end{align}
where $w(t)$ is a coefficient calculated with the scheduling parameter.

In classifier free guidance \cite{cfg}, the strength of the condition $y$ to the generated images can be controlled by changing the noise prediction as the following equation:
\begin{align}
    \tilde{\epsilon}_\phi (\bm{x}_t;y,t) = & ~ \epsilon_\phi(\bm{x}_t;\varnothing, t) \nonumber \\
    &+ s\cdot(\epsilon_\phi(\bm{x}_t;y,t)-\epsilon_\phi(\bm{x}_t;\varnothing,t))
\end{align}
where $\varnothing$ is a fixed null value.
The value of $s$ corresponds to the strength of $y$. By using larger $s$, we can generate images that are more faithful to the condition $y$.

\begin{figure*}
\begin{center}
\includegraphics[width=\textwidth]{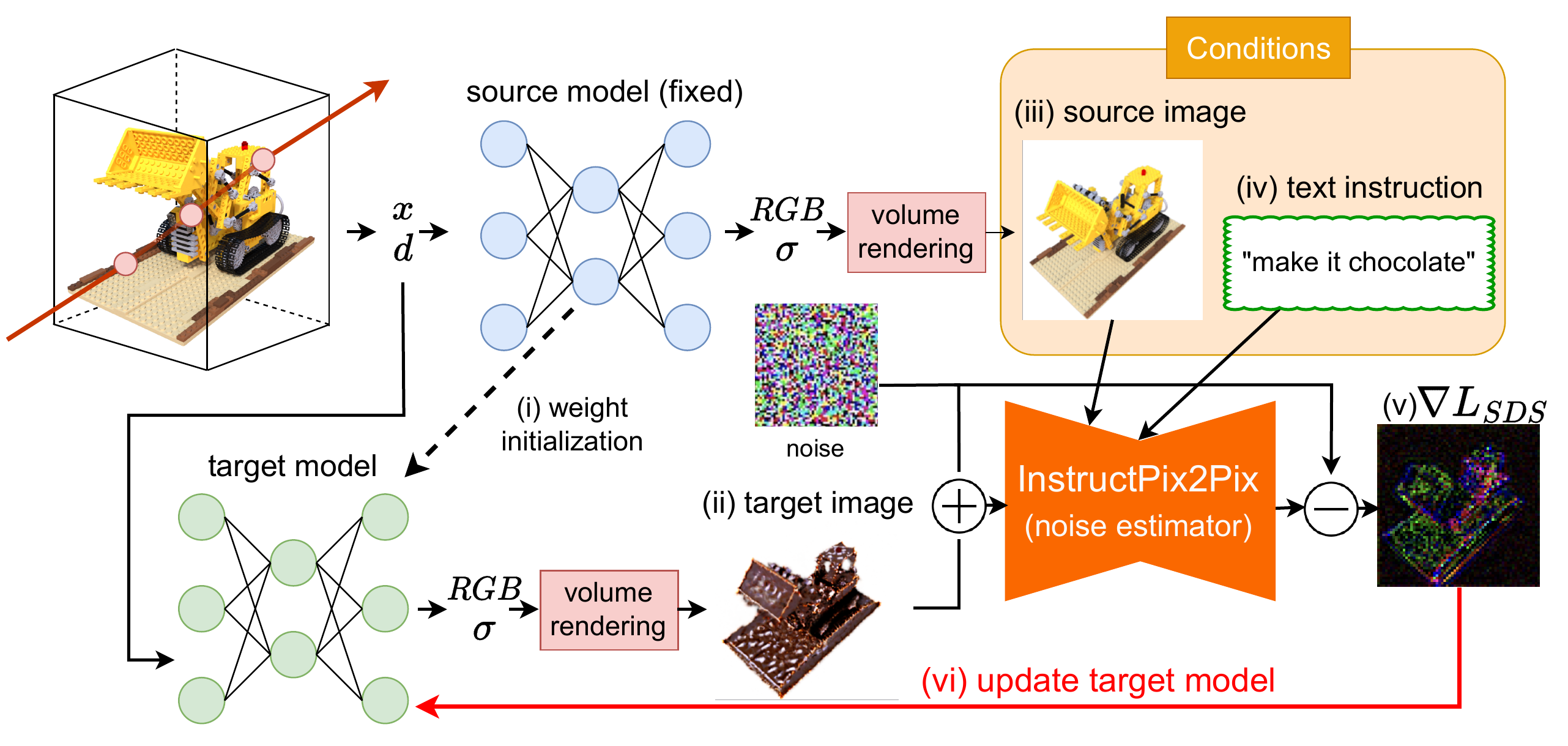}
\end{center}
   \caption{Overview of Instruct 3D-to-3D. First, the target model is initialized with the source model (i). Next, the target image is rendered from a random camera viewpoint (ii) and then the noise is added to input into InstructPix2Pix. The source image is rendered from the same viewpoint (iii) and input to InstructPix2Pix as conditions along with the text instruction (iv).  $\nabla\mathcal{L}_{\mathrm{SDS}}$ is calculated using them (v) and the target model is updated with it. By performing this procedure from various camera viewpoints, we can convert the target model along with the text instruction.}
\label{fig:overview}
\end{figure*}

\subsection{Text-guided Image-to-Image using Diffusion Models}
Diffusion models have also been extensively studied for text-guided Image-to-Image tasks. These methods can be divided into two categories based on what the input text represents.　
The first category assumes that the text describes the caption of the images before and after editing such as \cite{imagic, unitune, cyclediffusion, prompt2prompt}.
These methods require extra information not directly related to editing, such as descriptions of the part of the image that remains constant through editing.
On the other hand, the second category assumes that the text describes the instruction on what point should be changed through editing. InstructPix2Pix \cite{instructpix2pix} edits the image based on a given text instruction, which has the advantage of making image editing easier and more intuitive. 
They first create a dataset of editing instructions and image descriptions before and after editing with GPT-3 \cite{gpt3}. Then, they input it to \cite{prompt2prompt} to generate edited images. After that, they finetune StableDiffusion to generate these edited images conditioned by the source images and text instructions. Thus, 
InstructPix2Pix learns to edit images according to various text instructions.

In this paper, we extend InstructPix2Pix to 3D data and realize the task of editing a 3D scene into a new 3D scene with text instructions. This enables highly intuitive editing of 3D scenes.

\subsection{Implicit 3D Representation}
There are various ways to represent 3D information \cite{voxel2,achlioptas2018learning,brock2016generative,kanazawa2018learning,occupancy,nerf}. In this paper, we adopt implicit representation for its high expressive power, which has become quite popular since being used in NeRF \cite{nerf}. 
Given images that capture a target scene from different viewpoints, NeRF builds a neural network that predicts color and density at any spatial point in the scene from its 3D coordinates. This model is trained so that each viewpoint image matches the image rendered from the corresponding viewpoint based on the model.
After training, NeRF implicitly acquires a 3D representation and we can obtain images seen from any camera viewpoint. 

Recently, voxel grid-based implicit 3D representations have also been studied. They retain color and density information in the form of voxel grids instead of neural networks \cite{dvgo,plenoxels,nsvf} and achieve much faster training.  
In this study, we use voxel grid-based DVGO \cite{dvgo} to represent 3D scenes for fast 3D-to-3D conversion.

\subsection{Stylization of 3D scenes}
3D stylization is a task to transform a source 3D scene into a new scene that has a different style while preserving the content of the original scene. In \cite{arf, mishra2022clip}, the style is specified by a reference image.
On the other hand, CLIP-NeRF \cite{clipnerf} accepts texts to specify the style, which provides substantially higher flexibility. They finetune NeRF of the original scene so that the images rendered from any viewpoint would have high similarity in the CLIP feature space with the given text.
There is also a concurrent work \cite{nerfart} to ours for text-guided 3D stylization, which calculates a CLIP-based contrastive loss based on the source and target image to properly strengthen the 3D stylization. 

These methods only match the CLIP features of each viewpoint image and the reference image or text and do not use a generative model. Hence, there is no guarantee that they can convert the input 3D scene into a high-quality one. In this paper, we use a diffusion model to maximize the likelihood of each viewpoint image. In addition, while previous studies required a description of the converted 3D scene, we use editing instructions as input to make the 3D-to-3D conversion more intuitive.

\subsection{Text-to-3D models}
DreamFields \cite{dreamfields} was the first study to realize Text-to-3D. DreamFields generates a 3D scene that follows the input text from any viewpoint by optimizing the CLIP features of each NeRF viewpoint image to match the input text. However, since DreamFields only optimizes the CLIP features and does not use a generative model, it may lead to generating unrealistic scenes that just cheat the similarity at the CLIP feature space.

DreamFusion \cite{dreamfusion} is the first method to apply diffusion models to the Text-to-3D task. DreamFusion uses pretrained Imagen \cite{imagen} to generate 3D scenes by optimizing each NeRF viewpoint image $\bm{x}$ to follow the input text. Specifically, they first apply noise $\bm{\epsilon} \sim \mathcal{N}(0, \bm{I})$ to the viewpoint image $\bm{x}$ according to the randomly sampled $t$, and obtain a noisy image $\bm{x}_t = \sqrt{\bar{\alpha}_t}\bm{x}+\sqrt{1-\bar{\alpha}_t}\bm{\epsilon}$.
This noisy image is used to calculate the gradient of the loss function $\nabla_{\theta} \mathcal{L}_{\mathrm{SDS}}$ as the following equation:

\begin{align}
    \nabla_{\theta} \mathcal{L}_{\mathrm{SDS}} = \mathbb{E}_{t,\epsilon} \left[ w(t)(\epsilon_\phi (\bm{x}_t ; y, t) - \bm{\epsilon}) \frac{\partial \bm{x}}{\partial \theta} \right]
\end{align}
where $\theta$ is the parameters of NeRF and $y$ is the text description of the 3D scene to be generated. $\theta$ is updated using this gradient from any viewpoint. This method enables the generation of high-quality 3D scenes for a variety of text inputs.
They also edit generated 3D scenes by re-training them with new texts.
However, direct finetuning of a 3D scene may result in a scene that is far removed from the original 3D scene. In addition, this method requires a text description of the scene after conversion, and conversion by text instructions is not possible.

\begin{algorithm}[t]
\caption{Proposed method: Instructed 3D-to-3D}
\label{alg:algorithm1}
\textbf{Input}:\\
    \qquad $y$: instruction text,\\
    \qquad $\theta_\mathrm{src}$: parameters of source model,\\
    \qquad $g_\theta$: volume rendering function from a model parameterized with $\theta$\\
\textbf{Output}:\\
    \qquad $\theta_\mathrm{tgt}$: parameters of target model\\
\begin{algorithmic}[1] 
\State $\theta_\mathrm{tgt} \leftarrow \theta_\mathrm{src}$ \# initialize $\theta_\mathrm{tgt}$
\For{$i=1$ to $N_{\mathrm{iter}}$}
    \State \# rendering from source \& target model
	\State $c = \mathrm{random\_camera\_pose}()$
    \State $I_\mathrm{src} = g_{\theta_\mathrm{src}}(c)$
    \State $I_\mathrm{tgt} = g_{\theta_\mathrm{tgt}}(c)$
    \State $L_\mathrm{tgt} = \mathrm{StableDiffusionEncoder}(I_\mathrm{tgt})$
    \State \# add noise
    \State $\bm{\epsilon} \sim \mathcal{N}(0, \bm{I})$
    \State $t \sim U[1,\ldots, T]$
    \State $\bm{x}_t = \sqrt{\bar{\alpha}_t}L_\mathrm{tgt}+\sqrt{1-\bar{\alpha}_t}\bm{\epsilon}$
    \State \# calculate the gradient of the loss function
    \State $\nabla_{\theta_\mathrm{tgt}} \mathcal{L}_{\mathrm{SDS}} = \mathbb{E}_{t,\epsilon} \left[ w(t)( \tilde{\epsilon}_\phi (\bm{x}_t; y, I_\mathrm{src}, t) - \bm{\epsilon}) \frac{\partial L_\mathrm{tgt}}{\partial \theta_\mathrm{tgt}} \right]$
    \State $\theta_\mathrm{tgt} \leftarrow \mathrm{Adam}(\theta_\mathrm{tgt}, \nabla_{\theta_\mathrm{tgt}} \mathcal{L}_{\mathrm{SDS}})$
\EndFor
\end{algorithmic}
\end{algorithm}

\section{Proposed Method}
\subsection{Pipeline of Instruct 3D-to-3D}
Figure \ref{fig:overview} shows the overview of Instruct 3D-to-3D. 
Our proposed method converts a source model, which is an implicit representation of a source 3D scene, into a new target model along with the text instruction. 

The main idea of our Instruct 3D-to-3D is to learn the target model from an arbitrary viewpoint using InstructPix2Pix conditioned by the source scene and the text instruction.
First, the target model is initialized with the source model. Next, using the target model, a target image $I_\mathrm{tgt}$ is rendered from a random camera viewpoint and is fed into the encoder of StableDiffusion to obtain the corresponding latent feature $L_\mathrm{tgt}$. We add noise $\bm{\epsilon} \sim \mathcal{N}(0, \bm{I})$ to it to make noisy latent $\bm{x}_t = \sqrt{\bar{\alpha}_t}L_\mathrm{tgt}+\sqrt{1-\bar{\alpha}_t}\bm{\epsilon}$. A source image $I_\mathrm{src}$ is also rendered from the same viewpoint as the target image using the source model. $\bm{x}_t$ is input to InstructPix2Pix along with the source image $I_\mathrm{src}$ and the text instruction $y$ as conditions. As InstructPix2Pix has two conditions, $I_\mathrm{src}$ and $y$, the noise is estimated as follows.
\begin{align}
    \tilde{\epsilon}_\phi (\bm{x}_t; y, I_\mathrm{src}, t) &= ~ \epsilon_\phi(\bm{x}_t;\varnothing, \varnothing, t) \nonumber \\
    &+ s_I \cdot(\epsilon_\phi(\bm{x}_t;\varnothing, I_\mathrm{src}, t) -\epsilon_\phi(\bm{x}_t;\varnothing,\varnothing,t)) \nonumber \\
    &+ s_T \cdot(\epsilon_\phi(\bm{x}_t; y, I_\mathrm{src}, t) -\epsilon_\phi(\bm{x}_t;\varnothing,I_\mathrm{src},t))
    \label{eq:cfg_proposed}
\end{align}
where $s_I$ is a hyper-parameter that determines the degree of fidelity to the information of the source image, and $s_T$ is a hyper-parameter that determines the degree of fidelity to the text instruction. 
In this way, our proposed method explicitly incorporates the source image and the text instruction.
Similar to DreamFusion, we update the target model by the following gradient.
\begin{align}
    \nabla_{\theta_\mathrm{tgt}} \mathcal{L}_{\mathrm{SDS}} = \mathbb{E}_{t,\epsilon} \left[ w(t)( \tilde{\epsilon}_\phi (\bm{x}_t; y, I_\mathrm{src}, t) - \bm{\epsilon}) \frac{\partial L_\mathrm{tgt}}{\partial \theta_\mathrm{tgt}} \right]
\end{align}
where $\theta_\mathrm{tgt}$ is the parameters of the target model, $\phi$ is the parameters of InstructPix2Pix, and　$w(t)$ is a scheduling coefficient and we set it as $1-\bar{\alpha}_t$ in the experiments.

By repeating this procedure from randomly chosen camera viewpoints, an arbitrary viewpoint image of the target model becomes the image appropriately converted from the same viewpoint image of the source model.
As a result, we can obtain a target 3D scene converted from the source 3D scene along with the text instruction.
\begin{figure}
\begin{center}
\includegraphics[width=\linewidth]{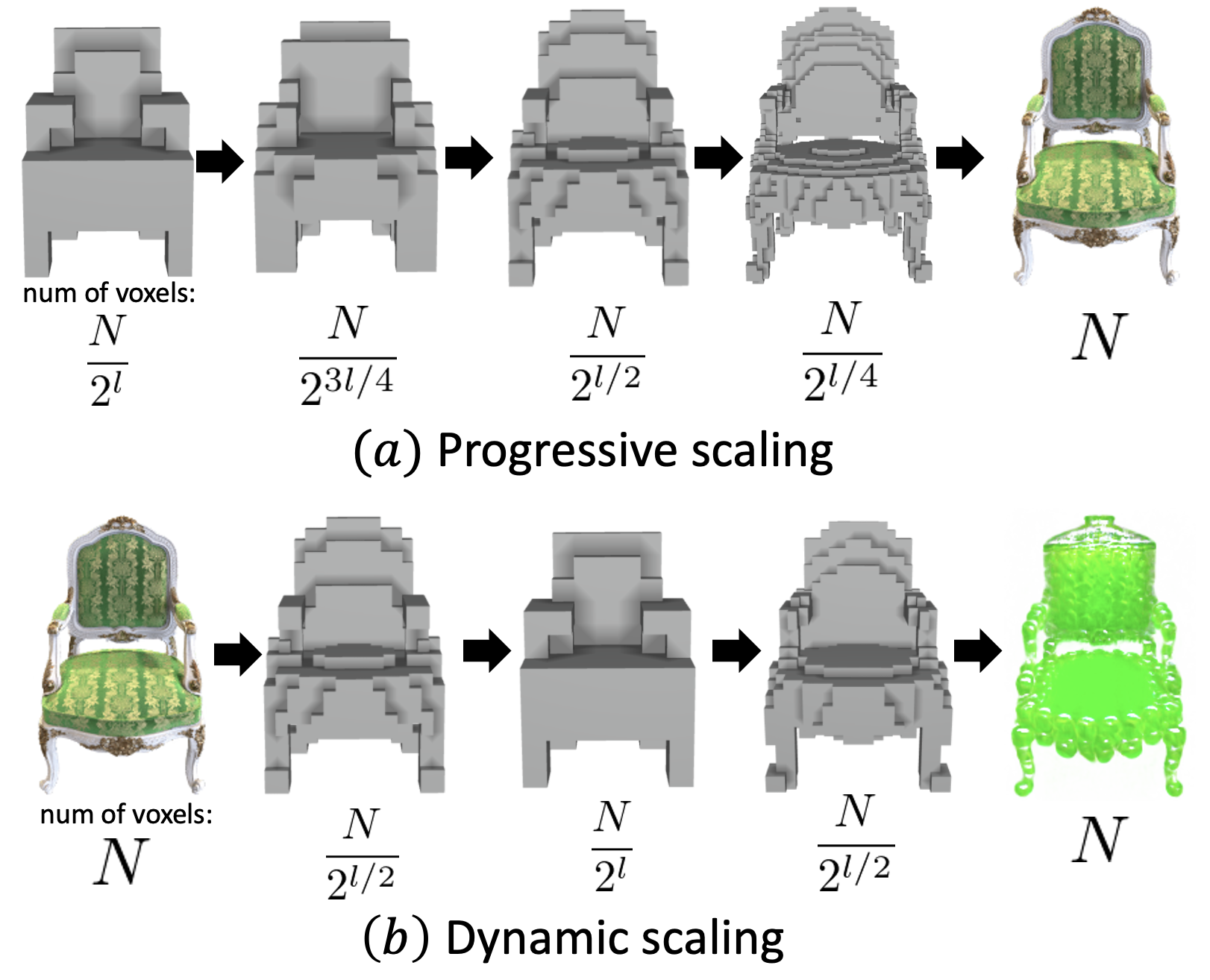}
\end{center}
   \caption{Comparison of progressive scaling and proposed dynamic scaling. In dynamic scaling, we first gradually decrease the number of voxels to change the global structure, then increase them to change to the local structure.}
\label{fig:dynamicscaling}
\end{figure}

\subsection{Dynamic Scaling}
In this study, we use DVGO \cite{dvgo} to perform fast 3D-to-3D conversions. DVGO is one of the voxel grid-based implicit 3D representations and  maintains density and color information in the form of a 3D voxel grid. The voxel grid is a discrete partition of 3D space, with each vertex holding color and density information. Volume rendering is performed with the interpolated information of the vertices around the ray.

The resolution of the 3D scene is determined by the number of voxels used in the model. DVGO performs progressive scaling \cite{nsvf} to gradually increase the number of voxels during training as shown in Figure \ref{fig:dynamicscaling} (a). It encourages the model to learn the rough geometry of the scene first, and then gradually move on to learning the fine details of the scene.

In our 3D-to-3D task, the number of voxels in the target model is initially set to $N$, the same as in the source model. In this situation, the voxel grid is too fine and it is difficult to change the geometry, and only the appearance changes. Therefore, we propose dynamic scaling, which gradually reduces the number of voxels from $N$ to $N/2^l$ during the 3D-to-3D conversion process and then gradually returns it to $N$. Figure \ref{fig:dynamicscaling} (b) shows this process. This allows gradual changes in the global structure, followed by changes in the detailed structure accordingly. Also, we can adjust the magnitude of the structure transformation by adjusting the scaling factor $l$. For large $l$, the number of voxels becomes small, which allows for a large structural transformation. Conversely, for smaller $l$, the structural transformation is suppressed.

\begin{figure*}
\begin{center}
\includegraphics[height=0.98\textheight]{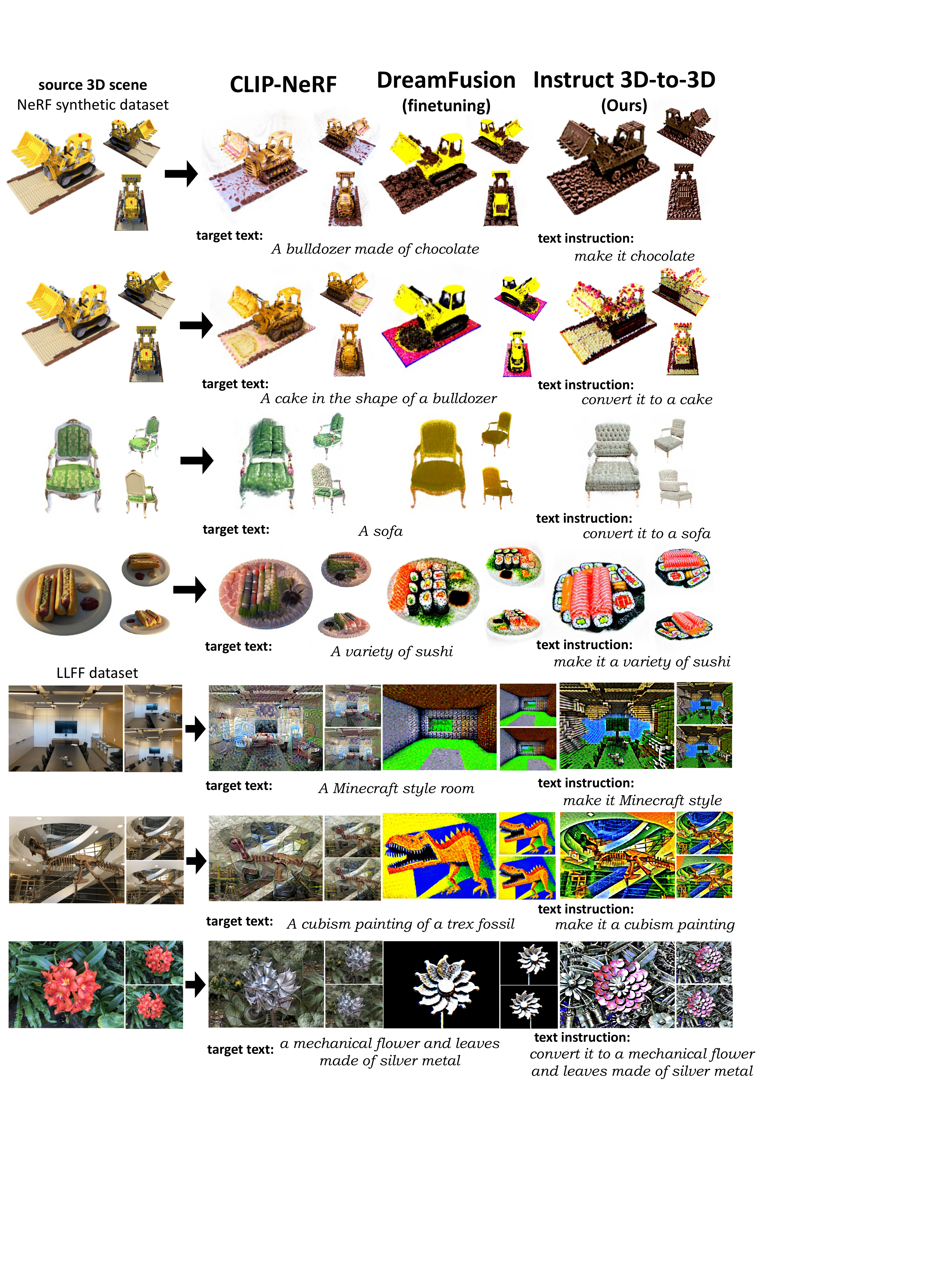}
\end{center}
   \caption{Qualitative comparison between the proposed method and baselines.}
\label{fig:qualitative}
\end{figure*}

\section{Experiments}

\subsection{Experimental Settings}
We implemented the proposed Instruct 3D-to-3D with PyTorch \cite{pytorch} and performed 3D-to-3D conversion with NeRF synthetic dataset \cite{nerf} and Local Light Field Fusion (LLFF) dataset \cite{llff}. 
We resized the NeRF synthetic dataset to 256$\times$256 and the LLFF dataset to 378$\times$504 for training.
In both datasets, we first trained the source models and then converted them with Instruct 3D-to-3D. The text instructions were manually designed.
The source and target models are constructed with DVGO and the number of voxels $N$ is set as $1,024,000$.

In the 3D conversion process, we trained the target model for 2,000 iterations. We set the dynamic scaling factor $l=4$. With dynamic scaling, we reduced the voxel number by a factor of $1/2^{l/5}$ every 150 iterations for the first 750 iterations and increased it by $2^{l/5}$ every 150 iterations for the next 750 iterations. After that, we kept the voxel number the same and trained for the remaining 500 iterations. Our proposed method can be completed in 15 minutes with a single NVIDIA A100 GPU. 

We set the text guidance scale $s_T$ of Eq. \ref{eq:cfg_proposed} to 100 according to \cite{dreamfusion}. The image guidance scale $s_I$ was set to $5$ for the NeRF synthetic dataset and $100$ for the LLFF dataset. We used the larger $s_I$ for the LLFF dataset to strongly maintain 3D consistency since it has only front-view images.

We also used CLIP-NeRF and DreamFusion as our baseline methods. Both were set up with the same experimental settings as the proposed method except for the loss function and the target texts. The target texts were also manually designed to match what the text instructions used in Instruct 3D-to-3D pointed to.

When fine-tuning source 3D scenes with DreamFusion, we used open-source StableDiffusion \cite{sd} instead of Imagen \cite{imagen}.

\begin{figure}
\begin{center}
\includegraphics[width=0.9\linewidth]{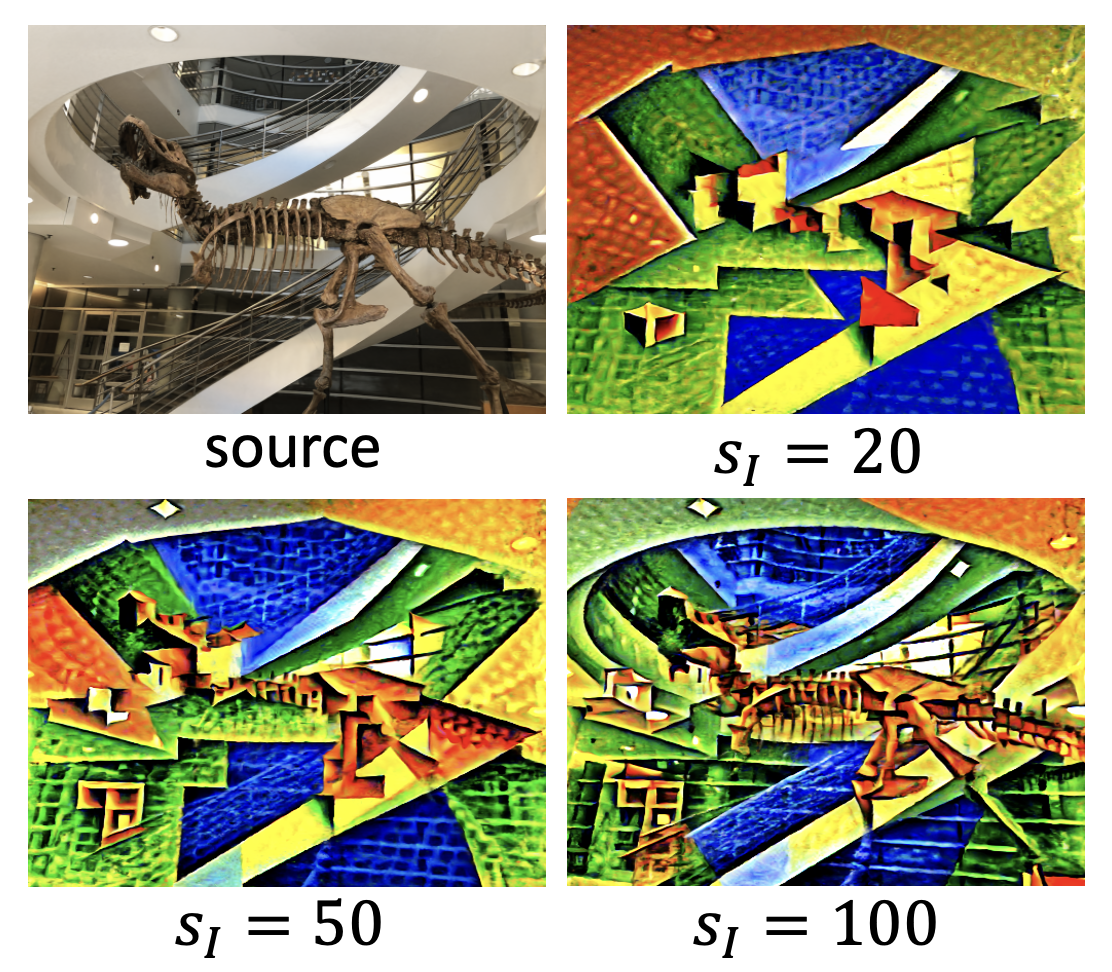}
\end{center}
   \caption{Effects of $s_I$ in the converted 3D scenes. We can preserve the structure of the source scene with larger $s_I$.}
\label{fig:image_guidance_scale}
\end{figure}

\subsection{Qualitative Evaluations}
\label{sec:qualitative}
We show the qualitative comparison between the proposed method and baseline methods in Figure \ref{fig:qualitative}. The results in the top four rows of Figure \ref{fig:qualitative} are examples from the NeRF synthetic dataset and results in the bottom three rows are examples from the LLFF dataset.
The text instructions used in Instruct 3D-to-3D and the target texts used in DreamFusion and CLIP-NeRF are also shown in Figure \ref{fig:qualitative}.
As a whole, it can be seen that the proposed method can produce converted 3D scenes that accurately reflect both the text instructions and structures of the source 3D scenes.
Although CLIP-NeRF can convert 3D scenes from the NeRF synthetic dataset relatively cleanly, it cannot convert 3D scenes from the LLFF dataset well, resulting in noisy 3D scenes. In addition, DreamFusion does not explicitly incorporate source 3D scenes as conditions, so it converts the LLFF 3D scenes into completely different 3D scenes.

Figure \ref{fig:image_guidance_scale} shows the differences by changing the image guidance scale $s_I$ in Eq. \ref{eq:cfg_proposed}. We can manipulate the degree to which the structure of the source 3D scene is reflected by adjusting the value of $s_I$. For large $s_I$, the source image condition is strongly incorporated during the noise estimation and the structure of the source 3D scene is strongly reflected.

\begin{figure}[t]
  \begin{minipage}{0.49\linewidth}
    \centering
    \includegraphics[width=\linewidth]{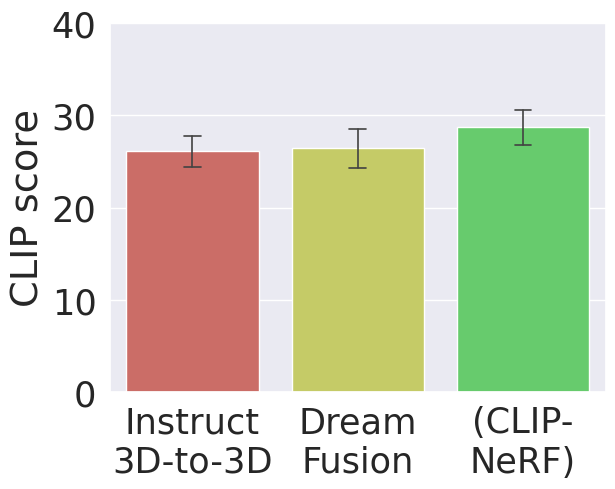}
    \caption{Average of CLIP scores (higher is better).}
    \label{fig:clip}
  \end{minipage}
  \begin{minipage}{0.49\linewidth}
    \centering
    \includegraphics[width=\linewidth]{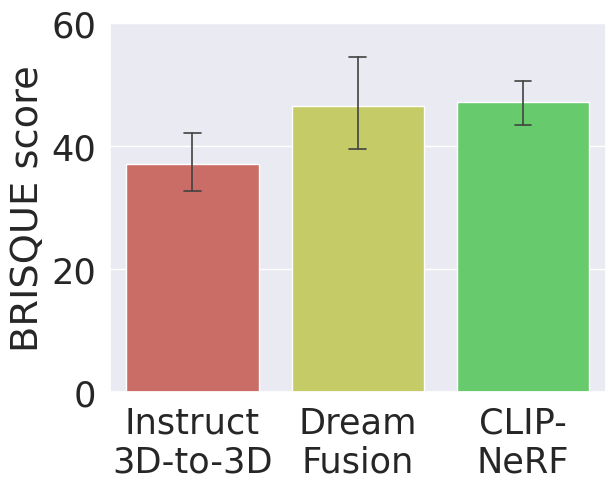}
    \caption{Average of BRISQUE scores (\textbf{lower is better}).}
    \label{fig:brisque}
  \end{minipage}
\end{figure}

\subsection{Quantitative Evaluations}
\label{sec:quantitative}
For quantitative evaluation, we measured CLIP score \cite{clip} and BRISQUE score \cite{mittal2012no}. The CLIP score is a measure of the semantic alignment of image-text pairs, where higher is better. The BRISQUE score is a measure of image quality, where lower means better.

We performed 3D-to-3D conversion by creating ten 3D scene-text pairs for each of the proposed and baseline methods.　The input texts were manually designed so that the target texts used in DreamFusion and CLIP-NeRF match what the text instructions in Instruct 3D-to-3D point to.　The list of scene-text pairs used in this experiment is shown in the supplementary material.
For each converted 3D scene, we rendered images from 100 viewpoints and used them to measure scores.　We did this for 10 converted 3D scenes, for a total of 1,000 rendered images used in the evaluation.
The CLIP score was calculated using the rendered images and the target texts, while the BRISQUE score was calculated from the rendered images only.

Figure \ref{fig:brisque} shows the BRISQUE score measurement results, and Figure \ref{fig:clip} shows the CLIP score measurement results.
Regarding the BRISQUE score, Instruct 3D-to-3D performs better than DreamFusion and CLIP-NeRF and achieves higher-quality 3D-to-3D conversion than these baseline methods.
On the other hand, the CLIP scores for Instruct 3D-to-3D and DreamFusion are comparable.
Although CLIP-NeRF achieves the best CLIP score, we cannot compare CLIP-NeRF with the other methods in terms of CLIP score in a fair manner. This is because CLIP-NeRF directly uses CLIP for training to improve the CLIP score. 
From these results, we confirmed that Instruct 3D-to-3D is able to convert high-quality 3D scenes with high text fidelity comparable to DreamFusion. The reason for this is considered to be that Instruct 3D-to-3D simultaneously incorporates the source 3D scene as a condition in addition to the text instruction, which contributes to a natural 3D scene conversion.

\subsection{User Study}
\label{sec:userstudy}
\begin{figure}[t]
  \begin{minipage}{0.49\linewidth}
    \centering
    \includegraphics[width=\linewidth]{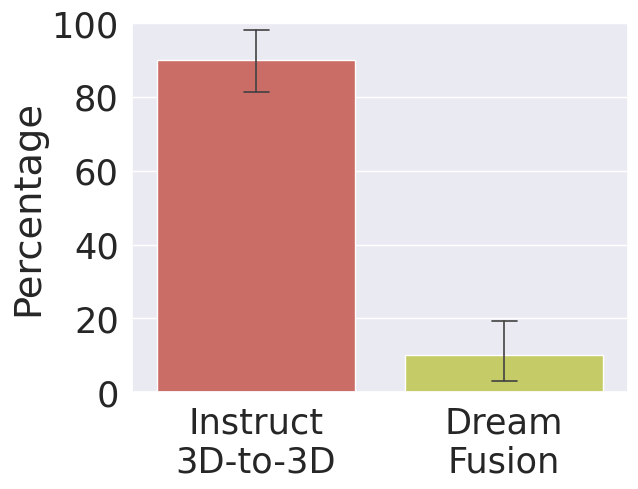}
  \end{minipage}
  \begin{minipage}{0.49\linewidth}
    \centering
    \includegraphics[width=\linewidth]{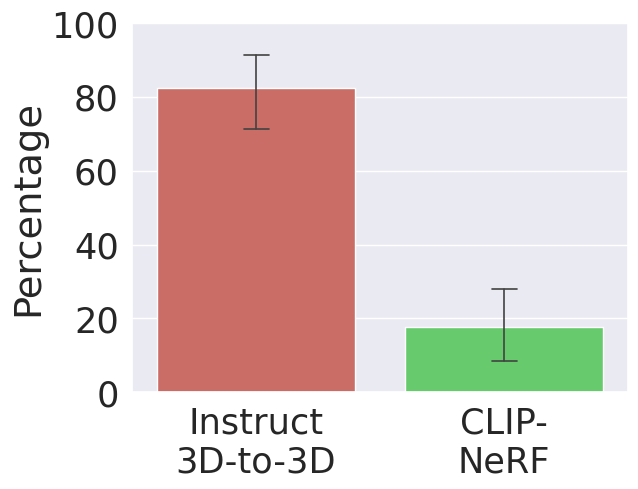}
  \end{minipage}
  \caption{User study results. Preference rates for 3D conversion quality of Instruct 3D-to-3D over DreamFusion and CLIP-NeRF.}
\label{fig:userstudy}
\end{figure}

To evaluate the perceptual quality of the converted 3D scenes, we conducted a user study.
We used the results of the 10 patterns of 3D-to-3D conversions used in the section \ref{sec:quantitative} as test cases for the user study.
In each test case, we first showed the source 3D scene as a video and the text instruction. Then we showed two of each of the three 3D scenes transformed with the proposed method and baseline methods in random order. The participants chose the better converted 3D scene by jointly considering the following four aspects: image quality from any viewpoint, 3D consistency, fidelity to the text instruction, and fidelity to the source 3D scene. We did not set a time limit. We finally collected 25 questionnaires. Figure \ref{fig:userstudy} shows the results of the user study. Our proposed method outperforms the baseline methods with a much higher user preference rate.

\begin{figure}
\begin{center}
\includegraphics[width=\linewidth]{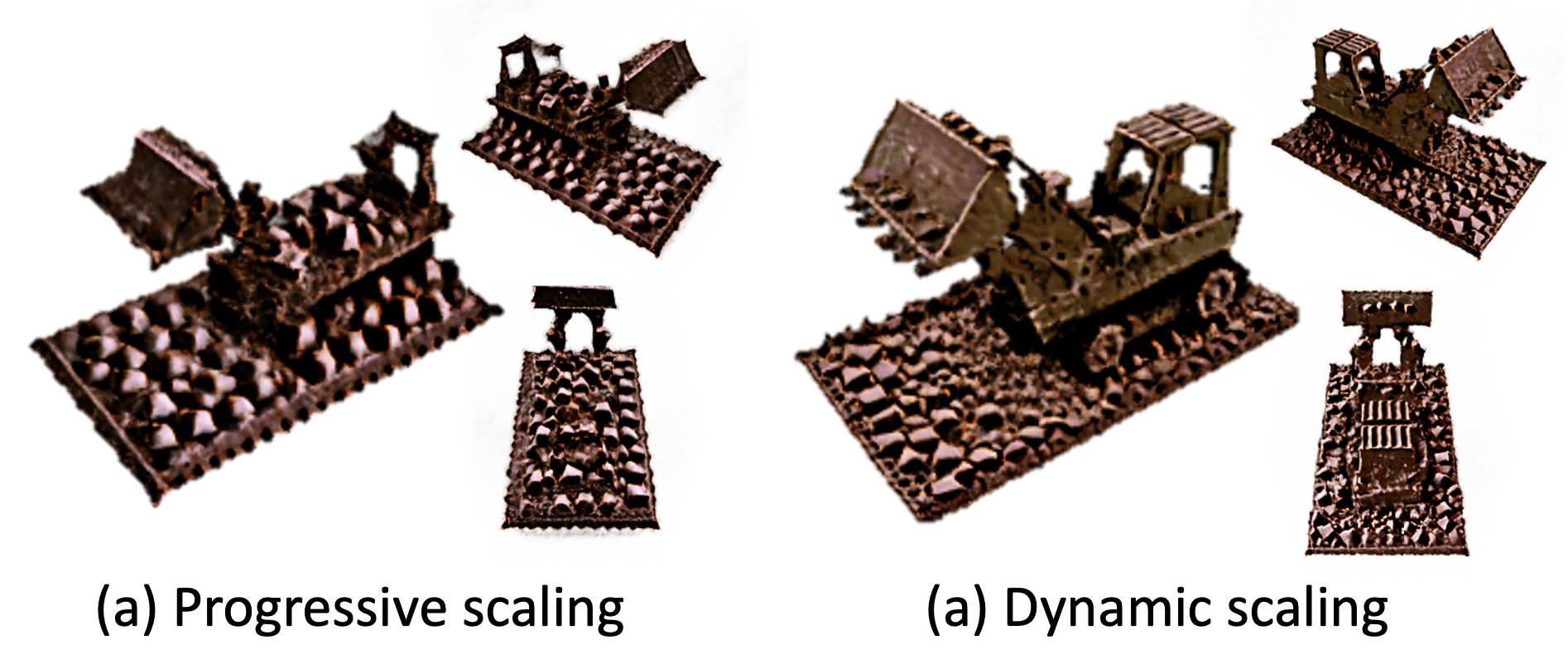}
\end{center}
   \caption{Comparison of 3D conversion results by scaling method. our dynamic scaling does not break the 3D structure.}
\label{fig:scaling_comparison}
\end{figure}

\subsection{Sensitivity to the Scaling Strategy}
We proposed dynamic scaling to gradually change the 3D structure. However, it is also possible to use conventional progressive scaling in 3D-to-3D conversion. Figure \ref{fig:scaling_comparison} shows a comparison of 3D conversion results using progressive scaling and dynamic scaling. As progressive scaling greatly reduces the number of voxels at the beginning of the conversion, the detailed 3D structure cannot be maintained. Our dynamic scaling gradually changes the number of voxels, resulting in small and repairable 3D structural damage and preserving the 3D structure.

\begin{figure}
\begin{center}
\includegraphics[width=0.9\linewidth]{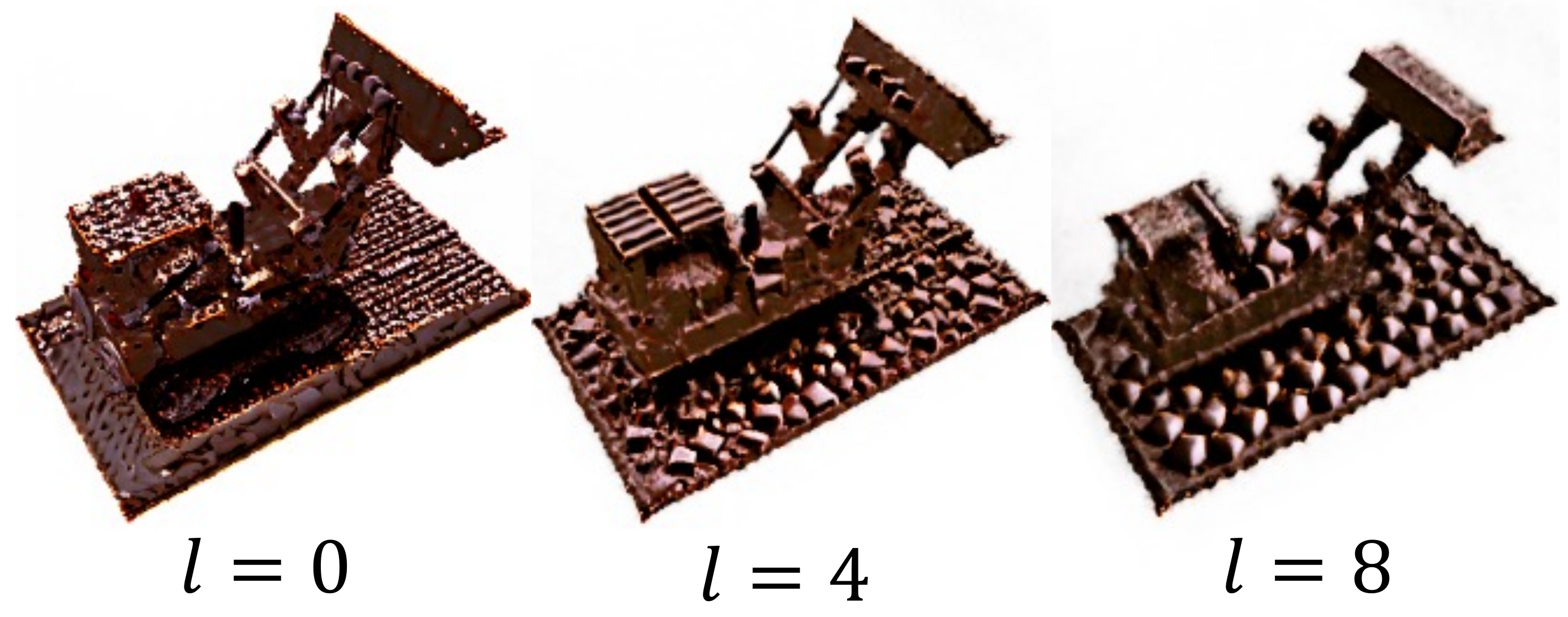}
\end{center}
   \caption{Effects of dynamic scaling factor $l$ in the 3D conversion. The large $l$ causes a major change in structure.}
\label{fig:effect_scaling}
\end{figure}

Figure \ref{fig:effect_scaling} shows the differences in 3D-to-3D conversion results by changing the dynamic scaling factor $l$. 
For larger $l$, the resolution of the voxel becomes smaller during the conversion process and the structure can be changed significantly.

\begin{figure}
\begin{center}
\includegraphics[width=0.85\linewidth]{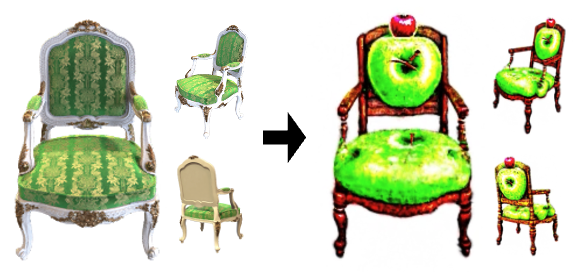}
\end{center}
   \caption{A failure case of 3D-to-3D conversion. This 3D scene is converted from a 3D scene of a chair with the text instruction "put an apple on the chair".}
\label{fig:putapple}
\end{figure}

\subsection{Limitations}
Figure \ref{fig:putapple} is an example of a failure case. We gave the instruction to place an apple on the chair, but the seat and backrest of the chair changed to apples. Similar to InstructPix2Pix, our proposed method is difficult to follow instructions that require spatial reasonings.
To solve this problem, it is necessary to handle spatial information, for example, by taking the depth information of the 3D scene into account. We leave this improvement to future work.

\section{Conclusion}
In this study, we proposed Instruct 3D-to-3D, which achieves a high-quality 3D-to-3D conversion following the text instruction.
We proposed dynamic scaling, which allows manipulation of the strength of the 3D structure transformation and smooth 3D-to-3D conversion. We performed quantitative and qualitative evaluations and showed that our Instruct 3D-to-3D outperforms the baseline methods in terms of the quality of the converted 3D scenes and the fidelity to the source 3D scenes and text instructions. Our proposed method makes 3D content easier to edit and use, and will contribute to greatly expanding the scope of various content productions.
\newpage
{\small
\bibliographystyle{ieee_fullname}
\bibliography{egbib}
}

\end{CJK}
\end{document}